\documentclass[lettersize,journal]{IEEEtran}
\usepackage{amsmath,amsfonts}
\usepackage{algorithmic}
\usepackage{algorithm}
\usepackage{array}
\usepackage[caption=false,font=normalsize,labelfont=sf,textfont=sf]{subfig}
\usepackage{textcomp}
\usepackage{stfloats}
\usepackage{url}
\usepackage{verbatim}
\usepackage{graphicx}
\usepackage{cite}
\usepackage[dvipsnames]{xcolor}
\usepackage{soul}

\begin{document}

\title{Astrocyte Regulated Neuromorphic Central Pattern Generator Control of Legged Robotic Locomotion}


\author{Zhuangyu Han, Abhronil Sengupta,~\IEEEmembership{Senior Member,~IEEE}
\thanks{Z. Han, and A. Sengupta are with the School of Electrical Engineering and Computer Science, The Pennsylvania State University, University Park, PA 16802, USA. E-mail: sengupta@psu.edu.}
}



\maketitle

\begin{abstract}
Neuromorphic computing systems, where information is transmitted through action potentials in a bio-plausible fashion, is gaining increasing interest due to its promise of low-power event-driven computing. Application of neuromorphic computing in robotic locomotion research have largely focused on Central Pattern Generators (CPGs) for bionics robotic control algorithms - inspired from neural circuits governing the collaboration of the limb muscles in animal movement. Implementation of artificial CPGs on neuromorphic hardware platforms can potentially enable adaptive and energy-efficient edge robotics applications in resource constrained environments. However, underlying rewiring mechanisms in CPG for gait emergence process is not well understood. This work addresses the missing gap in literature pertaining to CPG plasticity and underscores the critical homeostatic functionality of astrocytes - a cellular component in the brain that is believed to play a major role in multiple brain functions. This paper introduces an astrocyte regulated Spiking Neural Network (SNN)-based CPG for learning locomotion gait through Reward-Modulated STDP for quadruped robots, where the astrocytes help build inhibitory connections among the artificial motor neurons in different limbs. The SNN-based CPG is simulated on a multi-object physics simulation platform resulting in the emergence of a trotting gait while running the robot on flat ground. $23.3\times$ computational power savings is observed in comparison to a state-of-the-art reinforcement learning based robot control algorithm. Such a neuroscience-algorithm co-design approach can potentially enable a quantum leap in the functionality of neuromorphic systems incorporating glial cell functionality.
\end{abstract}


\section{Introduction}
\IEEEPARstart{C}{urrent} brain-inspired engineered systems have primarily focused on the emulation of bio-plausible computational models of neurons and synapses. Incorporation of other cellular units from the brain is lacking. Over the past few years, a growing body of evidence has demonstrated that glial cells, and in particular astrocytes, play an important role in the maintenance and modulation of neuronal dynamics facilitating brain function \cite{allam2012computational}. This work is inspired by recent theories suggesting that rich temporal dynamics such as synchrony form the dynamical basis of learning and memory and are expected to play a key role in enabling a \textit{dynamical view of intelligence} for the design of next-generation brain-inspired engineered learning systems (see report based on 6th US/NIH BRAIN Initiative Investigators Meeting \cite{monaco2021brain}). In particular, our work focuses on spinal central pattern generators (CPGs), which are neural circuits that generate spontaneous rhythmic patterns and serve as a model ``brain-like" system to design engineered learning platforms, specifically robotic locomotion controllers.

Multiple parallel works have pointed toward the critical role of astrocytes in spinal CPGs. Recent literature has reported that astrocytes in the spinal cord can influence the activity of the spinal motor network. Specifically, the excitation of spinal cord astrocytes causes a decline in fictive locomotion frequency \cite{acton2017gliotransmission}. Neuroscientists have further illustrated that excited astrocytes suppress the rhythmic locomotion activity by releasing ATP, which is metabolized into adenosine (ADO) immediately. Adenosine works as an antagonist of the excitatory neurotransmitter dopamine (DA) in the motor neurons \cite{acton2018modulation}. Also, researchers have observed that, in larval zebrafish, the futile swimming effort can activate radial astrocytes, which suppresses continuous unsuccessful attempts \cite{mu2019glia}. Further, it is reported that such inhibitory effect of astrocytes on swimming in zebrafish is mediated by a signaling pathway involving ATP - adenosine - $\text{A}_2$ receptor \cite{chen2025norepinephrine}. Such results are consistent with the claim that astrocytes play a homeostatic role in the neural system \cite{verkhratsky2018physiology}. Driven by such insights, we investigate the homeostatic regulation role of astrocytes in neuromorphic Spiking Neural Network (SNN) based CPG circuits for gait learning in quadruped robots. Neuromorphic robotic control has been recently gaining attention due to the promise of enabling low-power edge robotics in resource-constrained environments \cite{rostro2015cpg, polykretis2020astrocyte, gutierrez2020neuropod}. However, the works remain preliminary with application to simple robotic platforms primarily due to overtly simplistic computational modeling of the biological system, such as single-neuron rhythm generators. This work forges stronger connections with neuroscience computational models of CPG and astrocyte regulated synaptic plasticity to develop an algorithmic learning framework for gait emergence in legged robotic locomotion. The key distinguishing factors of our work against prior proposals are:

\textbf{(i) Neuroscience Inspired CPG Formulation:} Based on the bio-inspiration premise that CPG formulations need to have stronger neuroscience correlation in order to harness the efficiency of biological systems, we design a CPG circuit that captures the temporal dynamics of the locomotion unit and inter-limb communication to a much higher degree of detail. As shown in the paper, the bio-inspiration route enables us to scale neuromorphic robotic control, for the first time, to quadruped systems.

\textbf{(ii) Astrocyte Regulated CPG Plasticity:}  While the dynamics of structures mediating CPGs have been extensively studied \cite{grillner2021cpgs,ausborn2021computational}, rewiring or plasticity of CPGs remains relatively underexplored \cite{righetti2006dynamic,jouaiti2018hebbian}, especially in terms of understanding the homeostatic role of astrocytes. This work illustrates that astrocytes, working in parallel with local learning rules like reward-modulated Spike Timing Dependent Plasticity (STDP), suppress over-activated motor neurons directly or indirectly, by creating inhibitory synapses. This enables the emergence of a gait that provides continuous and smooth locomotion, while allowing us to harness the benefits of local, event-driven learning capability of bio-inspired systems when implemented on neuromorphic chips - paving the pathway for edge processing in robotics \cite{sandamirskaya2022rethinking}.

Additionally, the proposed algorithm is neuromorphic hardware compatible. For instance, stochastic leaky-integrate-fire spiking neuron characteristics used herein can be emulated by magnetic tunneling junction spintronic devices \cite{sengupta2016magnetic}. Additionally, the astrocyte models considered in this work are all local models, which contain only local variables and communicate with only the neighboring components in the architecture. As a result, the proposed algorithm is both bio-inspired and compatible with existing neuromorphic hardware or can be implemented by simple extensions of current neuromorphic computing architectures \cite{davies2021advancing}. 

\section{Related Work and Main Contributions}
In recent years, global policy optimization for robotic locomotion through reinforcement learning has been widely investigated \cite{haarnoja2018learning, tan2018sim, fu2021minimizing}, where a policy network receives a system state vector including the joint positions and velocities, and outputs the driving signal to the motors. The reinforcement learning alleviates the difficulty of tuning complex parameter sets. However, the policy network-driven locomotion usually requires continuous neural network inference for every control time step, which is computationally expensive. Further, training such systems is even more computationally challenging, thereby limiting their applicability for on-chip learning in edge robotic systems with resource constraints. 

Another category of robotic locomotion algorithms involves CPG circuits organizing the collaboration among joints or limbs, where one CPG unit controls a single joint or multiple joints. Some literature has applied classical control methods, such as phase oscillators, for the CPG \cite{lewis2002gait, liu2009cpg, zhang2014trot}. From the neuromorphic implementation standpoint, there has been some preliminary work aimed at constructing the CPG by using SNNs \cite{rostro2015cpg, polykretis2020astrocyte, gutierrez2020neuropod}. However, existing works on SNN CPG use over-simplified neuron models and network architectures. For example, the flexor and the extensor motor neuron pools are usually simplified into a network of two neurons, where the internal fine structure is neglected. The lack of inclusion of biological details in the CPG architecture has resulted in limited flexibility of the control system, thereby constraining their applicability to mostly simple robotic platforms like hexapod robot, in contrast to the more complex design space of real-world robotic locomotion control of more intensively researched models, such as quadruped robots. In this work, we forge stronger connections with theoretical neuroscience to develop a detailed bio-inspired CPG model and show that astrocyte control is instrumental to ensure optimal and stable gait emergence in robotic quadruped locomotion systems. Local learning mediated gait emergence ensures compatibility of our proposed control system with neuromorphic hardware, thereby leading to the potential of enabling real-time, low-power on-chip learning.

Finally, we also note that while recent work has explored the role of astrocytes in neuromorphic computing, it primarily remains limited to self-repair functionalities \cite{wade2012self, liu2017spanner,liu2018exploring,rastogi2021self,han2023astromorphic} and more recently self-attention mechanisms \cite{kozachkov2022building}, working memory \cite{tsybina2022astrocytes}, energy minimization in neural networks \cite{fountain2022effect}, structure learning \cite{han2023astronet}, and so on.

\section{Quadruped Locomotion Control Circuit}
\subsection{Quadruped System Overview}

The defining feature of a quadruped system (applicable to both biological systems and robots) is the set of four limbs. In the locomotion of a quadruped system, the contact between the ground and the feet of the limbs provides support to the torso and acceleration to the entire animal/robot. In case of a biological system, the lower motor neurons (located in the ventral horn of the spinal cord) directly control the tension of the joint flexors and extensors in the limbs \cite{monster1977isometric}. The dynamics of the joint movement in the gait during locomotion, specifically the torque, angular position, angular velocity and angular acceleration of the joints are determined by the activity of the lower motor neuron themselves and the communication among the lower motor neurons pools; between the upper motor neuron (located in the brainstem) and the lower motor neuron; and from the Golgi tendon organ to the lower motor neurons \cite{purves2018neuroscience, grillner2019current}. The gait locomotion functionality of a cat is still complete even when the cerebrums are removed, which indicates that the locomotion control system of four-limbed vertebrates is largely a local neural circuit in the brainstem and spinal cord \cite{whelan1996control}. Deriving insights from such neuroscience studies, we develop a bio-inspired CPG controller that models the collaboration among the limbs of a quadruped system. The framework is evaluated for a quadruped robot in a multi-body interaction physics simulation environment. In this paper, the Unitree A1 quadruped robot \cite{unitreea1} system is adopted. The dynamics of the robot is simulated in the MuJoCo platform \cite{todorov2012mujoco}.

\begin{figure*}[ht]
    \centering   
    \includegraphics[width=1\textwidth]{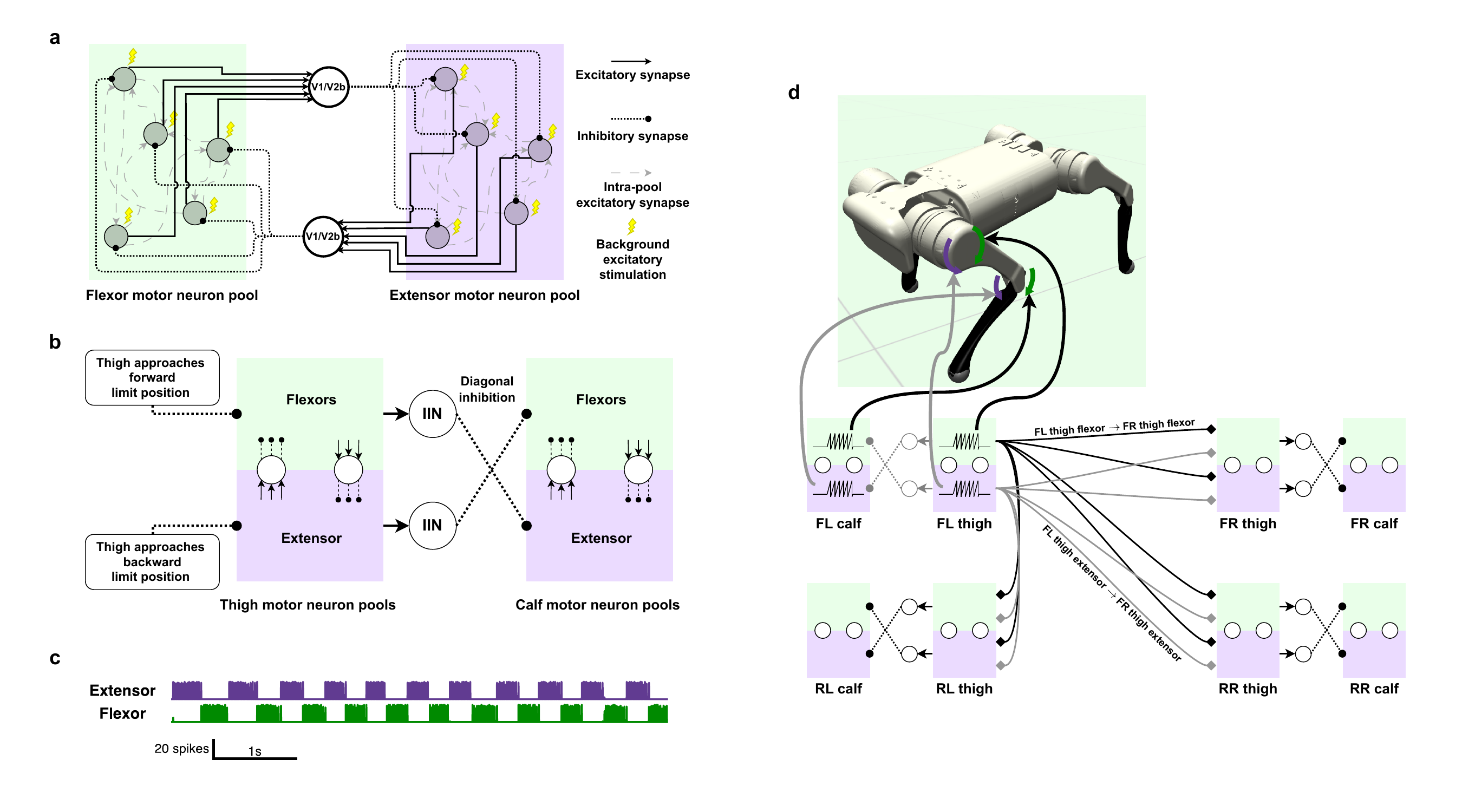}
    \caption{\textbf{SNN-based CPG architecture.} \textbf{a}, The reciprocal inhibition of flexor and extensor motor neuron pools, where the flexor and extensor motor neuron pools are connected through V1/V2b inhibitory interneurons. For clarity, only 5 motor neurons for each pool are displayed. The symbols for synapses in the following sub-figures are consistent with those in sub-figure \textbf{a}. \textbf{b}, The thigh muscle inhibits its antagonistic muscle in the calf through an inhibitory interneuron (IIN). \textbf{c}, A sample snapshot of spontaneous rhythmic burst pattern output from the reciprocal inhibition structure, under non-feedback condition. The vertical axis represents the number of motor neurons spiking at each time step. \textbf{d}, The motor neuron pool of the thigh sends synapses (solid lines with diamond ends) to the motor neuron pool of the thigh in all other limbs. It is worth noting here that these synapses could be either excitatory or inhibitory. For clarity, only the synapses from the front-left limb are displayed. FR: front-right, FL: front-left, RR: rear-right, RL: rear-left.}
    \label{architecture}
\end{figure*}

Figure \ref{architecture} shows the proposed CPG architecture for robotic locomotion, which can be summarized below. A locomotion unit for a single joint contains 2 motor neuron pools: the extensor and the flexor. Through internal connections, the extensor and flexor motor neurons can generate alternating bursts. The extensor and flexor activities are converted into the direction and magnitude of the joint actuator torque of the corresponding joint. Through STDP learning in the inter-limb connection modulated by the astrocyte and reward signal, thigh locomotion limbs inhibit or excite each other, through which the robot's gait emerges.

\subsection{Central Pattern Generator}

Critical behaviors of the gait locomotion of four-limbed animals are: (i) the alternating activity of the flexor and extensor muscles in the limbs \cite{grillner1975detailed, grillner2019current} and (ii) the phase difference among the limb movements. The alternating excitation of the two muscles of a single joint allows the joint to swing forward and backward periodically, which further converts the limb movement between the swing and stance phases \cite{collins1993coupled}. In the past decade, research has elucidated the alternating excitation mechanism of the flexor and extensor motor neuron pools -  reciprocal inhibition \cite{kiehn2016decoding}. Past research has divided the entire CPG for gait locomotion into locomotion units that individually control the movement of a single limb \cite{grillner2019current}. Within the locomotion units, the motor neurons to an individual muscle inhibit or excite the motor neurons to other muscles \cite{grillner2019current}. The phase difference among the limbs usually corresponds to the specific locomotion speed of the animal \cite{kiehn2016decoding}. Based on the phase difference, the gaits are usually categorized into walking, trotting and bounding \cite{kiehn2016decoding}. The following sections will explain (a) how a locomotion unit, controlling a single limb, is composed of reciprocally inhibiting spiking motor neuron groups and controls the joint movement, and (b) how the connections among all the locomotion units (i.e., inter-limb connection) determine the phase difference among the limbs.

\subsubsection{\textbf{Locomotion Unit}}
Related experiments illustrate that the flexor and extensor muscles of a single joint always excite in opposite phases, i.e., when the flexor muscle bursts, the extensor muscle does not fire, and vice versa \cite{grillner2019current}. Also, the duration of the burst of both the flexor and extensor is approximately constant, which indicates that an active muscle will spontaneously stop firing after a specific duration of time \cite{grillner2019current}. Such a burst pattern raises two critical questions: \textit{(1) How does the motor neuron burst terminate spontaneously after a specific duration of time?} (2) \textit{How does the halt of one muscle's activity trigger the activity of the antagonistic muscle?} We derive qualitative inspiration from neuroscience to answer these questions. The continuous firing of a motor neuron will accumulate $\mathrm{Ca^{2+}}$ in the cytoplasm, which later opens the $\mathrm{K^+}$ channels \cite{grillner2019current}. The decrement in the spike frequency of one muscle's motor neuron pool leads to the decline of inhibition to the antagonistic motor neuron pool. Based on this insight, in this paper, we assume that each motor neuron is receiving continuous background excitatory stimulation, and consequently, the flexor and the extensor reciprocally compete with each other, which finally forms periodic complementary burst patterns. The structure of such a reciprocal inhibiting SNN for a single joint is shown in Figure \ref{architecture}a. Two types of neurons are contained in this architecture: the motor neuron and the V1/V2b inhibitory interneuron. The motor neuron model is Pacemaker Stochastic Leaky-Integrate and Fire (PSLIF), which includes the cytoplasmic $\mathrm{Ca^{2+}}$ dynamics and the opening and closing of the $\mathrm{K^+}$ channel. The dynamics of the membrane potential of the motor neuron $i$ in the motor neuron pool can be formulated as:
\begin{equation}
    \begin{split}
        &\frac{d\,v_{\mathrm{motor},i}(t)}{dt} 
        = \frac{(v_{\mathrm{motor\_rest}}-v_{\mathrm{motor},i}(t))}{\tau_{\mathrm{motor}}}\\
        &+ I_{\mathrm{background}}(t,v)\cdot(1+\mathbf{Noise}(t)) \\
        & -C_{\mathrm{K\_chan}} \cdot \mathbf{Sigmoid}(S_{\mathrm{K\_chan}}\cdot([\mathrm{Ca}^{2+}] - \mathrm{Thres}_{\mathrm{Ca}^{2+}}))\\
        &+w_{\mathrm{V1/V2b}\xrightarrow{}\mathrm{motor}} \cdot \delta(t - t_{\mathrm{V1/V2b\_spike}})\\
        &+\sum_{\mathrm{j\neq i}}[w_{\mathrm{motor},j\xrightarrow{}i} \cdot \delta(t - t_{\mathrm{motor\_spike},j})]
    \end{split}
\end{equation}
Here, $v_{\mathrm{motor},i}$ is the membrane potential of neuron $i$ in the pool. $\tau_{\mathrm{motor}}$ is the motor neuron membrane potential decay time constant. $I_{\mathrm{background}}(t,v) = I_{\mathrm{background},0} + k_{\mathrm{background},v}\cdot v_{\mathrm{torso}}$ is the continuous background stimulation. $I_{\mathrm{background},0}$ is the strength of the background stimulation when the magnitude of the velocity of the torso is zero. $k_{\mathrm{background},v}$ is a positive background stimulation increment coefficient, and $v_{\mathrm{torso}}$ is the magnitude of the velocity of the torso. The velocity-related background stimulation will lead to an increasing frequency of the swing of the limbs when the robot is speeding up. The strength of the background stimulation also includes randomness. In the discrete implementation $\mathbf{Noise}(t)= \mathrm{A}_{\mathrm{random}}\cdot U_{[-1,1]}(t)$ where $U_{[-1,1]}(t)$ is a sampling of a random variable whose distribution is $\mathrm{Uniform[-1,1]}$. The random variable is sampled at every time step of the simulation. $\mathrm{A}_{\mathrm{random}}$ is the amplitude of randomness. $C_{\mathrm{K\_chan}}$ is the potential drop rate when the $\mathrm{K^+}$ channel is fully on. $\mathbf{Sigmoid}(x) = \frac{1}{1+e^{-x}}$. $S_{\mathrm{K\_chan}}$ is the sensitivity of the $\mathrm{K^+}$ channel to the $\mathrm{Ca^{2+}}$ concentration. $[\mathrm{Ca}^{2+}]$ is the cytoplasmic $\mathrm{Ca^{2+}}$ concentration. $\mathrm{Thres}_{\mathrm{Ca}^{2+}}$ is the $[\mathrm{Ca^{2+}}]$ threshold for opening the $\mathrm{K^+}$ channel. $w_{\mathrm{V1/V2b}\xrightarrow{}\mathrm{motor}}$ is a negative synapse strength from the V1/V2b interneuron to the motor neurons, and this value is uniformly the same for all the motor neurons. $w_{\mathrm{motor},j\xrightarrow{}i}$ is the intra-pool excitatory synapse strength from motor neuron $j$ to motor neuron $i$. $t_{\mathrm{V1/V2b\_spike}}$ and $t_{\mathrm{motor\_spike},j}$ are the spike timings of V1/V2b interneuron and motor neuron $j$ respectively. The cytoplasmic $\mathrm{Ca^{2+}}$ dynamics is:
\begin{equation}
    \frac{d[\mathrm{Ca^{2+}}]}{dt} = \frac{-[\mathrm{Ca^{2+}}]}{\tau_{\mathrm{Ca^{2+}}}} + r_{\mathrm{Ca^{2+}}} \delta(t-t_{\mathrm{spike}})
\end{equation}
where, $[\mathrm{Ca^{2+}}]$ is the concentration of cytoplasmic $\mathrm{Ca^{2+}}$, $\tau_{\mathrm{Ca^{2+}}}$ is the $\mathrm{Ca^{2+}}$ decay time constant, $r_{\mathrm{Ca^{2+}}}$ is the $[\mathrm{Ca^{2+}}]$ unit increment per action potential, $t_{\mathrm{spike}}$ is the time of firing of the motor neuron.
Also, the stochasticity of PSLIF neuron is attributed to the randomness of the condition to trigger an action potential. The probability of triggering an action potential by the PSLIF neuron at any time step is:
\begin{equation}
    p_{\mathrm{spike}} = \mathbf{Sigmoid}(\frac{v - v_{\mathrm{th}}}{S_{\mathrm{spike}}/2})
\end{equation}
where, $v$ is the membrane potential, $v_{\mathrm{th}}$ is the threshold potential, $S_{\mathrm{spike}}$ is transition width of spike probability. The motivation for introducing stochasticity is to prevent the simultaneous spike of all the motor neurons in a motor neuron pool.

For each muscle, the corresponding motor neuron pool contains 20 inter-connected motor neurons. The synapses among the motor neurons within a single pool are generated in a spatial manner: the location of each motor neuron is randomly generated in the 3-dimensional region $[0,1]\times[0,1]\times[0,1]$ from a uniform distribution and the synapses between a pair of motor neurons are generated according to the distance between two motor neurons, as shown below:
\begin{equation}
    w_{\mathrm{motor},j\xrightarrow{}i} = w_{0\mathrm{dist}} \cdot e^{-c_{\mathrm{dist}}\cdot \mathrm{dist}(j,i)}
\end{equation}
where, $w_{\mathrm{motor},j\xrightarrow{}i}$ is the synapse strength from motor neuron $a$ to motor neuron $b$, $w_{0\mathrm{dist}}$ is the synapse strength at zero distance, $\mathrm{dist}(j,i)$ is the spatial distance between motor neuron $j$ and motor neuron $i$, $c_{\mathrm{dist}}$ is the synapse strength spatial decay coefficient.

The inhibition is reported as a result of the activity of V1 and V2b inhibitory interneurons in the spinal cord \cite{zhang2014v1}. Thus, in our SNN design, every motor neuron pool excites a V1/V2b interneuron, which in turn inhibits the motor neuron pool of the antagonistic muscle. The neuron model of V1/V2b interneuron is the Stochastic Leaky-Integrate and Fire (SLIF), where the stochasticity is generated in a similar fashion as in PSLIF. The rhythmic pattern generated by the reciprocal inhibition is shown in Figure \ref{architecture}c.

While previous work has proposed the possible connection of motor neurons in one locomotion unit \cite{grillner2019current}, the mechanical structure of the limbs of the robot used in this paper is slightly different from a common quadruped animal, i.e., the number of joints responsible for locomotion in a single limb and the length ratio between the thigh and calf are different. Hence, the common quadruped locomotion unit architecture cannot be directly applied to the A1 robot. Therefore, we propose a locomotion unit SNN enabling the feet tips to move in a circular path (specific to the A1 robot). The diagram of the locomotion unit SNN is shown in Figure \ref{architecture}b. There is a second pair of motor neuron pools driving the flexor and extensor in the calf. Additionally, the thigh flexor inhibits the calf extensor, and vice versa. The neuron model of the inhibitory interneuron is SLIF, and they share the same model hyperparameters with the V1/V2b interneurons due to similar functionality. At the same time, the thigh flexor and extensors receive inhibitory sensory input when the thigh is approaching the limit position, which allows the transition between the swing and stance phases. For example, the angular range of the front thigh is $[0.6, 1.4]$. The front thigh flexor and the front extensor motor neurons will be inhibited when the thigh enters the regions $[0.6, 0.65]$ and $[1.35, 1.4]$ correspondingly. Given the diagonal inhibition from the motor neuron pools 
of the thigh to those of the calf, the phase transition of the calf is completely determined by the thigh motor neuron pools. Therefore the limit position inhibition for calf muscles is not necessary. 

Regarding driving the motors at robot joints, an angular impulse with the corresponding sign is generated every time the flexor and the extensor for this joint spike. The motor torque trace accumulates the angular impulses and decays exponentially when there is no stimulus. In the A1 robot, we use:
\begin{equation}
    \begin{split}
        \frac{d\,h_{\mathrm{thigh}}(t)}{dt} = &-\frac{h_{\mathrm{thigh}}(t)}{\tau_{h}} \\
        &+ I_{\mathrm{thigh}} \cdot \delta(t - t_{\mathrm{thigh\_extensor\_spike}}) \\
        & - I_{\mathrm{thigh}} \cdot \delta(t - t_{\mathrm{thigh\_flexor\_spike}})
    \end{split}
\end{equation}
where, $h_{\mathrm{thigh}}$ is the motor torque trace for the thigh joint, $\tau_{h}$ is the decay time constant of the motor torque trace, $I_{\mathrm{thigh}}$ is the increment of torque for a single action potential for the thigh joint and $t_{\mathrm{thigh\_extensor\_spike}}$ is the time of firing of any individual neuron in the thigh extensor motor neuron pool (similar discussions are valid for the flexor). It should be noted that, in a discrete implementation of the SNN simulation, multiple motor neurons can possibly (actually they frequently do) fire at the same time step. In such a case, all the concurrent spikes in this time step contribute to the torque. This discussion is valid for all the following scenarios where the time of firing of a motor neuron pool is considered. In the MuJoCo configuration of the robot provided by the manufacturer, the joint motors of the A1 robot are working in torque mode. Consequently, the value of $h_{\mathrm{thigh}}$ is directly provided as output to the thigh motor actuator. The calf joint motor is driven by the same dynamics, but the angular impulse, $I_{\mathrm{calf}}$, can be a different value.

\subsubsection{\textbf{Inter-limb Connection}}
Recent works have elaborated that there are excitatory and inhibitory synapses from the motor neurons for one limb to the motor neurons for another limb \cite{grillner2019current}. In this paper, the phase difference among the limbs is modulated by the inter-limb excitatory/inhibitory connections. As shown in Figure \ref{architecture}d, the thigh muscles from one limb send axons to all the muscles in all the other limbs. For example, the motor neurons of the thigh extensor in the front-left limb send connections to the thigh flexors and the extensors in front-right, rear-left and rear-right limbs. The calf motor neurons are not involved in the inter-limb connection. The source motor neuron pool and the target motor neuron pool are fully connected, but the entire connection shares a single synapse strength value. Such a connection could be expressed as a table of size 8 by 8, shown in Figure \ref{gait_connection}a, where the value in row $i$ and column $j$ represents the synapse weight of the connection projected from muscle $i$ to muscle $j$. The inter-limb connection is trained by reward-modulated Spike Timing Dependent Plasticity (STDP), including the necessary modulation of astrocytes, which will be introduced in a later section. In the next subsection, the cytoplasmic signaling pathway will be introduced. The computational detail regarding the inhibitory effect of astrocyte output on inter-limb connection will be illustrated in Section IV.

\subsection{Astrocyte Regulation}
The astrocyte is reported to play a homeostatic role in spinal cord locomotion control circuitry \cite{acton2017gliotransmission, mu2019glia, chen2025norepinephrine}. Specifically, the astrocytes, when activated, release signaling molecules that suppress over-excited neural groups to prevent energy waste. The spiking activity of neighboring neurons is a trigger of the cytoplasmic calcium activity of astrocytes.
In general, first, specific receptors on the plasma membrane of the astrocyte sense the product of the activated neighboring neurons, such as 2-arachidonyl glycerol (2-AG) or glutamate. Next, a sequence of interactions between $\mathrm{Ca}^{2+}$, inositol trisphosphate ($\mathrm{IP}_3$) and $\mathrm{IP}_3$ receptor responds to the activated membrane receptors. Finally, multiple types of signaling molecules are released from the astrocytes, such as glutamate or ATP, following the $\mathrm{Ca}^{2+}$ or $\mathrm{IP}_3$ activity.

Astrocytes are diverse across the entire nervous system. For the locomotion control system, a previous work suggested that Endocannabinoid receptor $\mathrm{CB}_1$ is expressed by the astrocytes in multiple brain regions and the dorsal horn of the spinal cord \cite{salio2002neuronal}. In this paper, we regard 2-AG, a type of endocannabinoid, as the effective neurotransmitter signaling from motor neurons to the astrocytes. The motor neurons release a constant amount of 2-AG every time there is an action potential. The concentration of 2-AG decays exponentially when no motor neuron fires:
\begin{equation}
    \frac{d([\mathrm{AG}])}{dt} = \frac{-[\mathrm{AG}]}{\tau_{\mathrm{AG}}} + r_{\mathrm{AG}}\delta(t-t_{\mathrm{motor\_spike}})
\end{equation}
where, [AG] is the concentration of 2-AG, $\tau_{\mathrm{AG}}$ is the decay time constant of 2-AG, $r_{\mathrm{AG}}$ is the production rate of 2-AG and $t_{\mathrm{motor\_spike}}$ is the time instant when the motor neuron spikes. 2-AG binds to $\mathrm{CB}_1$ receptor on the astrocyte and then triggers the signaling pathway inside the astrocyte. 

For the astrocyte-to-neuron signaling pathway, it is well established that ATP is the direct product of astrocytes that suppresses the locomotion activity \cite{acton2017gliotransmission, chen2025norepinephrine, acton2018modulation}. Nevertheless, the mathematical model for the relationship between the output signals of the astrocyte and the $\mathrm{Ca}^{2+}$ and $\mathrm{IP}_3$ activity in locomotion circuitry has not been detailed in prior works. Also, the detailed intra-astrocyte dynamics is absent in previous works on astrocyte-modulated locomotion. Instead, these works concentrate on the communication between astrocytes and the surrounding neurons \cite{acton2017gliotransmission, chen2025norepinephrine}. Therefore, a computational model for intra-astrocyte activity and ATP production needs to be adopted. A prior study \cite{gibson2007computational} proposed a single-astrocyte model whose output is extracellular ATP. However, according to the paper, the evidence supporting the relationship between $\mathrm{IP}_3$ and ATP release was not concrete \cite{gibson2007computational}. Chen \textit{et al.} also state that the calcium activity is believed to be the direct cause of the rise in extracellular ATP concentration \cite{chen2025norepinephrine}. Further, another work proposed a data-driven mathematical model of astrocyte, which, nevertheless, concentrates on the propagation of calcium activity in an astrocyte network \cite{hofer2002control}. As a result, adopting a customized model for describing the intra-astrocyte activity and the astrocyte-ATP relationship is inevitable. Here, in the absence of backing of in vivo experiments on astrocyte modulation in the spinal cord, we use the Li-Rinzel model for intra-astrocyte dynamics and use the unit ATP release model for the ATP production from astrocytes, where the ATP release directly depends on the astrocyte cytoplasmic $\mathrm{Ca}^{2+}$ activity. This is consistent with results reported in Refs. \cite{acton2017gliotransmission, chen2025norepinephrine} and at the same time minimizes the number of redundant assumptions. 

The Li-Rinzel model \cite{li1994equations}, which describes the general cytoplasmic $\mathrm{Ca}^{2+}$-$\mathrm{IP}_3$ dynamics, is the basis of many following astrocyte dynamics models \cite{manninen2018computational}. In the Li-Rinzel model, the activation of $\mathrm{CB}_1$ receptor results in $\mathrm{IP}_3$ release in the astrocyte cytoplasm. $\mathrm{IP}_3$ further regulates the release of $\mathrm{Ca}^{2+}$ from the Endoplasmic Reticulum (ER) \cite{wade2012self}. The transportation of $\mathrm{Ca}^{2+}$ between the cytoplasm and ER can be divided into three components: the channel current from ER to the cytoplasm, $J_{\mathrm{chan}}$; the passive leakage current from the ER to the cytoplasm, $J_{\mathrm{leak}}$; the active pumping current from the cytoplasm to ER, $J_{\mathrm{pump}}$. In summary:
\begin{equation} 
    \frac{d[\mathrm{Ca}^{2+}]}{dt} = J_{\mathrm{chan}} + J_{\mathrm{leak}} - J_{\mathrm{pump}} 
\end{equation}
The $[\mathrm{Ca}^{2+}]$ here is the concentration of cytoplasmic $\mathrm{Ca}^{2+}$. Readers seeking a detailed computational model are directed to \cite{wade2012self}.

\begin{figure*}[ht]
    \centering   \includegraphics[width=0.9\textwidth]{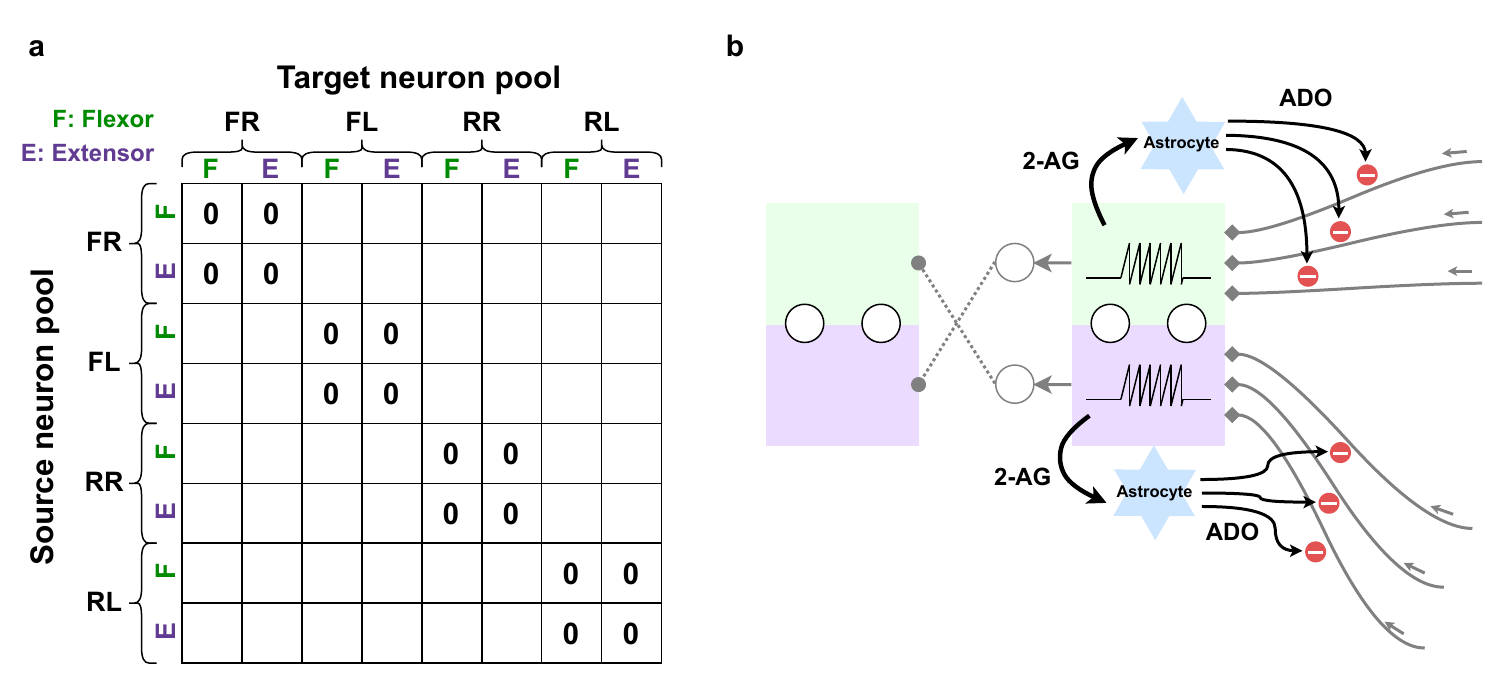}
    \caption{\textbf{Astrocyte regulated gait search.} \textbf{a}, The table containing the synapse weight of inter-limb connections. As there is no self-limb connection, the 2 by 2 sub-tables on the diagonal are all zero. \textbf{b}, The astrocyte receives 2-AG from the motor neuron pool, which is later activated and releases adenosine that decreases the strength of the input synapses to this motor neuron pool. For clarity, only three of the input synapses are displayed.}
    \label{gait_connection}
\end{figure*}

Chen \textit{et al.} indicate that the extracellular ATP rises following the calcium activity in astrocytes \cite{chen2025norepinephrine}. The extracellular ATP, once released, is metabolized into adenosine immediately \cite{acton2017gliotransmission, acton2018modulation, chen2025norepinephrine}. The adenosine plays an inhibitory role in the modulation of locomotion in the spinal cord \cite{acton2017gliotransmission, chen2025norepinephrine}. Multiple works have reported that the adenosine receptor on the membrane of the locomotion-related interneurons is a part of the inhibitory signaling pathway, but the specific type of receptor varies among different reports. Ref. \cite{acton2018modulation} suggests that adenosine activates the $\text{A}_1$ receptor of the locomotion-related neurons in mice, while another report demonstrates that $\text{A}_{2}$ is the target of adenosine in swimming behavior modulation in zebrafish \cite{chen2025norepinephrine}. As ATP is always metabolized into adenosine rapidly \cite{acton2017gliotransmission, acton2018modulation, chen2025norepinephrine}, and the time for degeneration from ATP to adenosine can be neglected, we simplify the astrocyte-ATP-adenosine dynamics into astrocyte-adenosine dynamics for reducing the complexity of the computational model. The product of ATP metabolism, adenosine, is regarded as the direct output of the astrocyte. It is also assumed that one unit of adenosine $r_{\mathrm{ADO}}$ is released every time when the calcium concentration exceeds a threshold. Then, adenosine diffuses in the extracellular space and experiences exponential decay.
\begin{equation}
    \frac{d[\mathrm{ADO}]}{dt} = \frac{-[\mathrm{ADO}]}{\tau_{\mathrm{ADO}}} + r_{\mathrm{ADO}}\delta(t-t_{\mathrm{Ca}^{2+},\mathrm{ADO\_prev}})
\end{equation}
Here, $[\mathrm{ADO}]$ is the concentration of adenosine in the nearby region of the astrocyte, $\tau_{\mathrm{ADO}}$ is the adenosine decay time constant, $r_{\mathrm{ADO}}$ is the production rate of adenosine, $t_{\mathrm{Ca}^{2+},\mathrm{ADO\_prev}}$ is the time instant when $[\mathrm{Ca}^{2+}]$ is above the threshold satisfying the condition that the time from the last adenosine release is not shorter than the adenosine refractory period. The adenosine received by the motor neurons negatively contributes to the synapse strength sent to this motor neuron pool. Figure \ref{gait_connection}b illustrates the negative feedback relationship between the motor neuron pool and the adenosine.

\section{Gait Learning Algorithm}
Previous works have discussed the reward-modulated STDP for encouraging actions through a delayed reward signal \cite{izhikevich2007solving}. The basic idea of gait search is to reward the gait, i.e., the phase difference of the bursts among all the motor neuron pools, providing a larger forward speed and more stable torso attitude. The effect of different collaborations of the muscles is usually delayed, i.e., a good gait leads to a large forward speed, but the speeding up procedure still costs a duration of time. Hence, reward-modulated STDP is necessary for gait search. The reward function is given by:
\begin{equation}
    \begin{split}
        r(t) &= \alpha_{\mathrm{vel}_x} \cdot \mathrm{vel}_x(t) \\
        &- \alpha_{\omega_{\mathrm{roll}}} \cdot |\omega_{\mathrm{roll}}(t)|\\
        &- \alpha_{\omega_{\mathrm{pitch}}} \cdot |\omega_{\mathrm{pitch}}(t)|\\
        &- \alpha_{\omega_{\mathrm{yaw}}} \cdot |\omega_{\mathrm{yaw}}(t)|
    \end{split}
\end{equation}
where, $\mathrm{vel}_x$ is the velocity of the torso in $x$ direction, which is the default heading direction at the beginning of the training process. $\omega_{\mathrm{roll}}$, $\omega_{\mathrm{pitch}}$ and $\omega_{\mathrm{yaw}}$ are the angular velocities in the roll, pitch and yaw directions. The parameters $\alpha_{\mathrm{vel}_x}$, $\alpha_{\omega_{\mathrm{roll}}}$, $\alpha_{\omega_{\mathrm{pitch}}}$ and $\alpha_{\omega_{\mathrm{yaw}}}$ are the reward coefficients. 
The reward-induced weight update from motor neuron pool $x$ to motor neuron pool $y$ in reward-modulated STDP used in this work can be written as:
\begin{equation}
    \begin{split}
        &\Delta w^\mathrm{reward}_{x,y}(t) =
        \eta_{\mathrm{eff}}(t)r_{\mathrm{eff}}(t)\cdot \mathrm{STDP}_{x,y}(t)\cdot \zeta_{w_{x,y}}(t)
    \end{split}
\end{equation}
where,
\begin{equation}
    \begin{split}
        \frac{d\,\mathrm{STDP}_{x,y}(t)}{dt} &= -\frac{\mathrm{STDP}_{x,y}(t)}{\tau_{\mathrm{STDP}}} \\
        &+\delta(t - t_{y\mathrm{\_spike}})\cdot u_{x}(t)\\
        &- \eta_{\mathrm{negative\_relative}}\cdot \delta(t - t_{x\mathrm{\_spike}})\cdot u_{y}(t)\\
        \frac{d\,u_{x}(t)}{dt} = &-\frac{u_{x}(t)}{\tau_{\mathrm{trace}}} + \delta(t - t_{x\mathrm{\_spike}})\\
        \eta_{\mathrm{eff}}(t) = &\eta\cdot \mathrm{Progress}(t)\\
        r_{\mathrm{eff}}(t) = &r(t) - c_{\mathrm{average}}\cdot\overline{r_{(t-0.1s, t)}}\\
        \zeta_{w_{x,y}}(t) = &\frac{[w_{\mathrm{max}} - w_{x,y}(t)][w_{x,y}(t) - w_{\mathrm{min}}]}{(w_{\mathrm{max}} - w_{\mathrm{min}})^2}
    \end{split}
\end{equation}
where, $\mathrm{STDP}_{x,y}(t)$ is the STDP signal from motor neuron pool $x$ to motor neuron pool $y$. $\tau_{\mathrm{STDP}}$ is the STDP signal decay time constant. $t_{y\mathrm{\_spike}}$ and $t_{x\mathrm{\_spike}}$ are the times of firing of motor neuron pool $y$ and 
$x$. $u_{x}(t)$ and $u_{y}(t)$ are the spike traces of motor neuron pool $x$ and 
$y$ at time $t$. $\tau_{\mathrm{trace}}$ is the spike trace decay time constant. $\eta_{\mathrm{negative\_relative}}$ is the relative negative learning rate, which controls the contribution of negative STDP signal in learning. $\eta_{\mathrm{eff}}(t)$ is the effective learning rate, which is the product of training progress $\mathrm{Progress}(t)$ and learning rate $\eta$. The definition of training progress can be found in the Experiments section.  $r_{\mathrm{eff}}(t)$ is the effective reward for STDP learning. $\overline{r_{(t-0.1s, t)}}$ is the sliding window average of the reward function from 0.1 seconds before to the current time. $c_{\mathrm{average}}$ is the averaged reward coefficient. By subtracting a fraction of $\overline{r_{(t-0.1s, t)}}$, the system encourages the concurrent firing of the muscles which provides a higher reward than the average. It also discourages the concurrent firing of the muscles leading to the falling of the robot, at which time the angular velocities of the torso are large. $\zeta_{w_{x,y}}(t)$ is a weight constraint factor preventing the weight from being out of the range of $[w_{\mathrm{min}}, w_{\mathrm{max}}]$.

It is worth noting here that the STDP learning algorithm is usually applied in situations where excitatory synapses play a significant role \cite{diehl2015unsupervised}. The learning rate for post-synaptic firing is usually much larger than its counterpart for pre-synaptic firing. As a result, the learning of inhibitory synapses, which are critical in many neural systems like ours, is largely unexplored. According to the previous section, adenosine released by the astrocytes is sensed by the neighboring neurons. For simplifying the computational model, the detailed interaction between the adenosine and motor neurons is neglected. It is assumed that the adenosine generates an inhibitory effect on the input synapses of the motor neurons. i.e., the adenosine reduces the weight of the inter-limb synapses from other motor neurons to the corresponding motor neuron pool, as illustrated in Figure \ref{gait_connection}b. Therefore, at every time step, the astrocyte-induced weight update from motor neuron pool $x$ to motor neuron pool $y$ is:
\begin{equation}
    \begin{split}
        &\Delta w^\mathrm{astrocyte}_{x,y}(t) =\\
        &-\eta_{\mathrm{ADO}}\cdot\mathrm{Progress}(t)\cdot[\mathrm{ADO}_y]\cdot\zeta_{w_{x,y}}(t)
    \end{split}
\end{equation}
where, $\eta_{\mathrm{ADO}}$ is the efficacy of adenosine in building inhibitory synapses, $\mathrm{Progress}(t)$ is the training progress, $[\mathrm{ADO}_y]$ is the adenosine concentration at motor neuron pool $y$, $\zeta_{w_{x,y}}(t)$ is the weight constraint factor. The complete weight update of the inter-limb connection is:
\begin{equation}
\label{eq_complete_weight_update}
    \Delta w_{x,y}(t) = \Delta w^\mathrm{reward}_{x,y}(t) + \Delta w^\mathrm{astrocyte}_{x,y}(t)
\end{equation}

For discussion on the computational modeling of the proposed system, please refer to the section \textit{Computational Implementation of the Dynamical System} in the supplementary material.

\section{Experiments}

\subsection{Problem Setup and Simulation Parameters}
The experiments are performed using a computational simulation framework containing the quadruped robot physics simulation and the discrete simulation of the SNN-based CPG. MuJoCo (https://github.com/deepmind/mujoco) version 2.1.0 is adopted in this paper for the quadruped robot simulation. mujoco-py \cite{mujocopy}, version 2.1.2.14, is used to call the MuJoCo functions through Python APIs. The MJCF model of the A1 robot is obtained from the manufacturer's repository (https://github.com/unitreerobotics/unitree\_mujoco). The simulations are run on a bare metal computer with one Intel(R) Core(TM) i9-13900K CPU, 64GBytes RAM and the Ubuntu 22.04.2 LTS operating system. To simulate the dynamics of the SNN, i.e., the neuron/astrocyte models and the synaptic learning rules, we implemented the model dynamics in Python in a discrete manner, without utilizing any external SNN library. The simulation frequency is 1000 Hz. The entire codebase of the SNN implementation can be found at \url{https://github.com/NeuroCompLab-psu/Astrocyte_Regulated_Gait_Search.git.}

The information about the A1 robot model configuration for simulation, model initialization and control parameters such as the coefficients of the PI controller can be found in the supplementary material.

The structure of the algorithm in a single time step can be split into three distinct steps: 
\begin{enumerate}
    \item SNN and astrocyte model status are updated. Motor torque output is calculated from motor neuron activities.
    \item MuJoCo simulates A1 model for one step
    \item Reward is calculated from robot sensor data. Inter-limb connection weight is updated.
\end{enumerate}

\subsection{Training Session and Training Progress}
The MuJoCo environment inherently can simulate the robot system for an indefinitely long time. However, based on the locomotion control architecture mentioned before, the robot cannot reset its own status after a fall. Therefore, splitting the entire duration of the simulation into training sessions is necessary. In this paper, a simulation session will terminate when either the time reaches the maximum session length or the ``non-alive counter'' of the robot has a value higher than a pre-defined threshold, which implies that the robot falls or the robot is stuck into a non-healthy attitude. The maximum session length is set to 10 seconds. The ``non-alive counter'' accumulator will increase by one for each time step the ``alive indicator'' is false. The ``alive indicator'' is a Boolean expression:
\begin{equation}
    \mathrm{ALIVE} = \mathrm{Z\_AXIS}_{\mathrm{torso}} \cdot (0,0,1) \geq \mathrm{Thres}_{z\_\mathrm{ALIVE}}
\end{equation}
where, $\mathrm{Z\_AXIS}_{\mathrm{torso}}$ is the torso's local Z-axis vector expressed in the world coordinate system. $\mathrm{Thres}_{z\_\mathrm{ALIVE}}$ is the threshold of the torso Z-axis vertical component. In an ideal still-standing attitude, the $\mathrm{Z\_AXIS}_{\mathrm{torso}}$ has a value $(0,0,1)$, which is vertically upward.

As the reward-modulated STDP reinforces the synapse weight when concurrent firing provides a positive reward, the synapse weights will continuously change even when a good gait allows the robot to run smoothly. Therefore, the learning rate should be dependent on the learning progress, in order to halt the learning when a well-performing gait has been captured. The training progress is defined as:
\begin{equation}
    \mathrm{Progress}(t) = \mathbf{Sigmoid}(-\frac{\frac{\overline{L_{[\mathrm{curr}-10,\mathrm{curr}-1]}}}{L_{\mathrm{max}}}-0.9}{0.02})
\end{equation}
where, $L_{[\mathrm{curr}-10,\mathrm{curr}-1]}$ is the average length of the previous 10 sessions. $L_{\mathrm{max}}$ is the maximum session length. Notice that the value of  $\mathrm{Progress}(t)$ is in the range of $[0,1]$. The training progress approaches zero when the average session length is approaching the maximum session length.

At the beginning of the entire training, the robot has a high possibility to fall to the ground. The training goal at the beginning, therefore, is learning not to fall. Progressing further, at the stage where the robot can run continuously for 7 to 8 seconds, the learning goal becomes maximizing the rewards. The beginning 2 seconds of the session, at this time, negatively contribute to the learning goal because of the zero resetting velocity. Consequently, we designed a dynamic learning period instead of entire-session learning. The starting time step of the learning period in any session is:
\begin{equation}
    \begin{split}
        &t_{\mathrm{learning\_start}}\\
        &=\mathrm{Clip}((\overline{L_{[\mathrm{curr}-10,\mathrm{curr}-1]}} - 1), 0, 2) \quad\mathrm{seconds}
    \end{split}
\end{equation}

\begin{figure*}[ht]
    \centering   
    \includegraphics[width=1\textwidth]{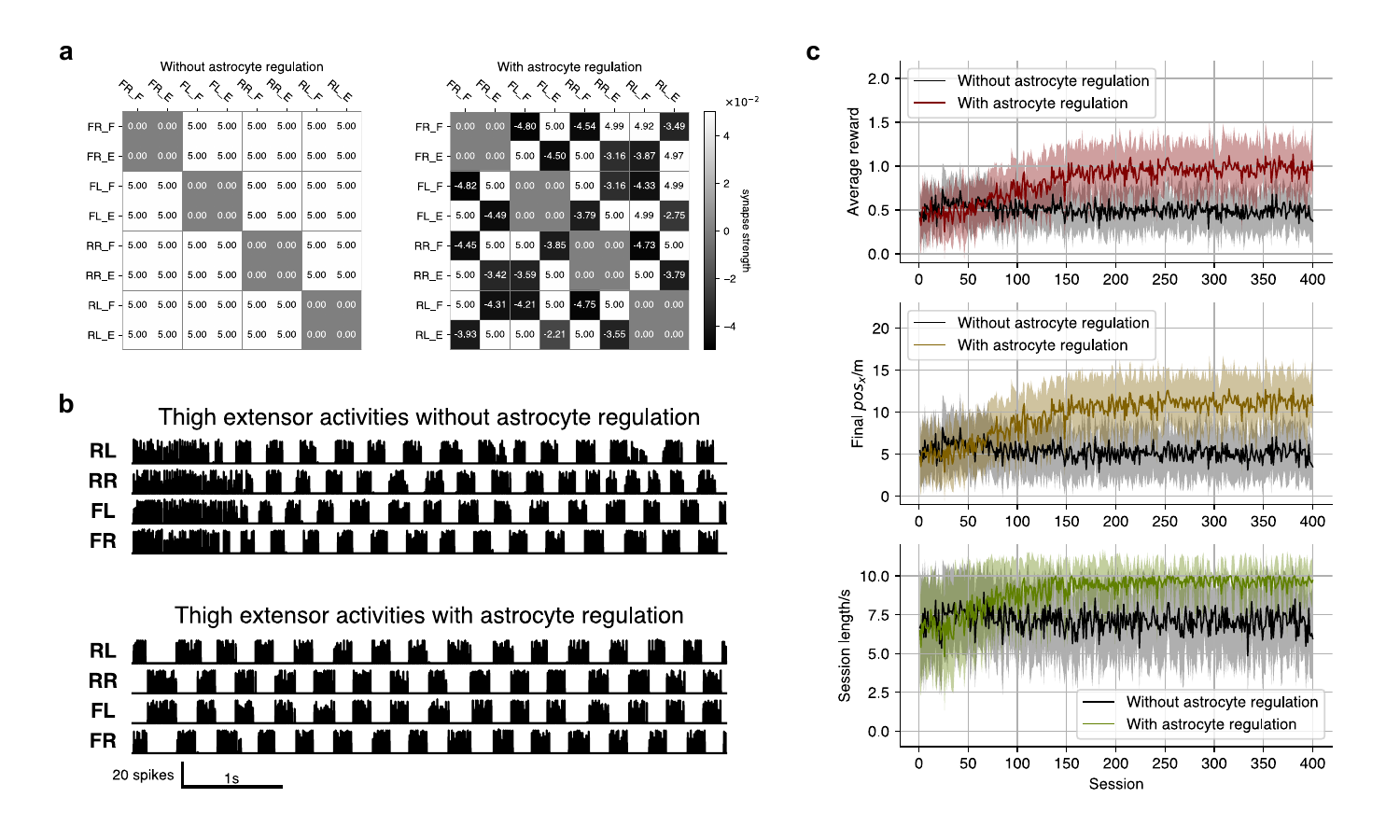}
    \caption{\textbf{Ablation studies.} \textbf{a}, Typical trained inter-limb connection weights. \textbf{b}, Typical thigh extensor activities of the four limbs after training. Upper panel: limb collaboration is absent when astrocyte regulation is removed. Lower panel: diagonal-synchronization emerges under astrocyte regulation. \textbf{c}, The statistics of the average reward, the displacement in $x$ direction at the final time step of each session, and session length in the training process, where the number of simulation trails is 20. The dark line represents the mean value, and the shaded region represents the interval of mean $\pm$ standard deviation.}
    \label{results}
\end{figure*}

\subsection{Gait Emergence and Ablation Study}
The entire training process can be divided into multiple training sessions. At the beginning of the training, all the synapses of the inter-limb connection have zero values. The limbs hence swing back and forth individually with random and time-variant phase differences but similar frequencies. In the process of such random phase difference generation, multiple types of collaboration among the limbs will temporally emerge. For example, trotting, pacing or bounding can be seen for short time periods, typically much shorter than the maximum session length, at the beginning stages of learning. Based on our observations in the performed experiments, such random swings usually lead to a fall or non-continuous slow running. Among all the emerged gaits, the optimal gait produces maximum $x$-directional speed and stable attitude of the torso. Nevertheless, the non-optimal gaits, generating slow locomotion, will also result in a positive reward, because of the large reward coefficient of the $x$-directional speed. The effect of the optimal and the non-optimal gaits all positively contribute to the inter-limb connections, through the STDP learning algorithm. Such always-positive learning causes saturated inter-limb synapse strengths at the pre-defined strength upper bound. Hence, there is a strong need to build inhibitory inter-limb connection weights, which is performed by the astrocytes through the release of adenosine to the motor neurons.

Figure \ref{results} shows the homeostatic contribution of the astrocytes. Removing the astrocyte regulation implies that the second term on the right-hand side of Equation \ref{eq_complete_weight_update} is removed. In Figure \ref{results}a, the left panel shows the trained inter-limb connection weights without the astrocyte model. All the synapses are trained into excitatory connections with saturated weight $w_{\mathrm{max}}$. In the right panel, it can be observed that the connection from one limb to another limb always contains two excitatory and two inhibitory synapses. For example, the connections from FR limb to FL limb have two inhibitory synapses in the diagonal positions and two excitatory connections in the anti-diagonal positions, which implies that the extensor of FR inhibits the extensor of FL (similar discussions are valid for the flexors). The extensor of FR excites the flexor of FL and vice versa. Such a combination produces a complementary firing pattern of the FR and the FL. In fact, the right panel of Figure \ref{results}a indicates a diagonal-synchronizing gait: FR-RL synchronized and FL-RR synchronized. The two groups are complementary to each other. This diagonal-synchronizing gait is usually called trotting. In summary, the astrocyte can filter out the effect of non-optimal gaits. Figure \ref{results}b more obviously illustrates the trotting gait. In the upper panel, the all-excitatory inter-limb connection brings unstable locomotion where the phase difference among the limbs is unstructured. Under current thigh and calf limit positions, $I_\mathrm{background}$, $I_{\mathrm{thigh}}$ and $I_{\mathrm{calf}}$, only trotting gait will emerge. For a real quadruped animal, for example, a cat, the muscle tension and CPG rhythmic pattern frequency increase along with the descending modulating signal from the brainstem. Therefore, for the emergence of other gaits, the above-mentioned parameters need to be specifically tuned for our robotic application scenario. Furthermore, this trotting gait is currently only applicable to flat ground. Adaptability to more complex terrain will be the scope of future work.
In Figure \ref{results}c, the average rewards in each session, the final $x$ coordinate and the session lengths are plotted. The final $x$ coordinate evaluates the distance the robot has moved as the reset heading direction is positive $x$ direction. The reward and moving distance in the case without astrocyte modulation fluctuate around a constant and do not significantly improve after training. In contrast, including astrocyte regulation improves the average reward and locomotion speed, because of the emergence of the optimal gait. More specifically, the average x-directional speed of the last 20 sessions is $1.17 \,\mathrm{m/s}$, which is close to the trotting gate speed manually set in prior literature \cite{fu2021minimizing}. Such a speed emerges naturally, and it is the result of specific CPG burst duration and motor torque hyper-parameters. A session length shorter than 10 seconds implies a fall of the robot in the session. Involving astrocyte regulation significantly decreased the probability of falling during running.

\subsection{Power Consumption Benefits}
Recent literature on conventional deep learning approaches for quadruped robot gait learning primarily involves reinforcement learning, where the action at each control time step is output from a policy network.
For example, \cite{fu2021minimizing} introduces an energy-related reward function in a reinforcement learning setting for quadruped (Unitree A1 robot) locomotion. The policy network is a 3-layer architecture where each hidden layer contains 128 neurons and the output layer contains 12 neurons, providing the target angular positions of the 12 joints in the limbs. The input to the policy network is a 42-dimensional state vector containing the current angular position and velocity of the joints in the limbs, the roll and pitch of the torso, the foot contact indicators and the action produced in the previous control time step.
Therefore, the power consumption of the control system can be regarded as the product of the energy consumption of a single feed-forward inference computation and the control frequency. For the feed-forward inference, the number of multiplication operations of a single layer $j$ with the bias of an artificial neural network is:
\begin{equation}
    \mathbf{Num}_{j,\mathrm{MULT}} = d_i\cdot d_o
\end{equation}
where, $d_i$ is the input dimension of this layer and $d_o$ is the output dimension of this layer. The number of addition operations of layer $j$ is:
\begin{equation}
    \mathbf{Num}_{j,\mathrm{ADD}} = [(d_i-1)+1]\cdot d_o = d_i\cdot d_o
\end{equation}
where $[(d_i - 1)+1]$ represents the number of additions, including bias, for a single output neuron. Therefore, the power consumption of the policy network based control system can be calculated as,
\begin{equation}
    \begin{split}
        P_{\mathrm{Policy}} = f_{\mathrm{control}} \cdot \sum_{j=1,2,3}(&\mathbf{Num}_{j,\mathrm{MULT}}\cdot\epsilon_{\mathrm{MULT}} \\
        + &\mathbf{Num}_{j,\mathrm{ADD}}\cdot\epsilon_{\mathrm{ADD}})
    \end{split}
\end{equation}
where, $f_{\mathrm{control}}=100$Hz is the control frequency \cite{fu2021minimizing}, $\epsilon_{\mathrm{MULT}}$ and $\epsilon_{\mathrm{ADD}}$ are the energy consumption of a single multiplication and a single addition operation. In this paper, we consider the multiplication and addition operations are implemented in $45$nm CMOS process \cite{han2015learning}, where $\epsilon_{\mathrm{MULT}} = 3.7\mathrm{pJ}$ and $\epsilon_{\mathrm{ADD}} = 0.9\mathrm{pJ}$. Consequently,
\begin{equation}
    \begin{split}
        P_{\mathrm{Policy}} = 100\mathrm{Hz} \cdot [&(42\cdot128)+(128\cdot128)+(128\cdot12)]\\
        &\times(3.7\mathrm{pJ}+0.9\mathrm{pJ})\\
        & = 1.07\times10^{-5} \,\, \mathrm{Watt}
    \end{split}
\end{equation}

The SNN-based CPG proposed in this paper operates by means of transmitting and receiving spikes. Instead of the multiplication and addition operation that takes place for each synaptic connection in a non-spiking network, only an ADD operation takes place per synaptic operation in an SNN upon the receipt of an incoming spike \cite{sengupta2019going}. The energy consumption of the SNN-based control system therefore can be calculated by counting the total number of accumulation operations. The discrete simulation of an SNN implies that the simulation frequency does not impact the firing frequencies of the spiking neurons. Thus, the power consumption of the SNN-based CPG is independent of the SNN simulation frequency, excluding the sensory input portion in this paper, which will be discussed in detail below. As elaborated in Section III, the neurons composing the CPG can be categorized into motor neurons and inhibitory interneurons, including the V1/V2b interneurons maintaining the reciprocal inhibition and the IINs between the thigh and the calf muscles in the same limb. These two types of inhibitory interneurons both individually inhibit 20 corresponding motor neurons, and hence they can be treated uniformly: one spike sent from an inhibitory interneuron results in 20 accumulation operations. One motor neuron in the calf flexor and extensor motor neuron pool excites 19 other motor neurons in the same motor neuron pool and excites a V1/V2b interneuron, which implies that one spike from a calf muscle motor neuron also leads to 20 accumulation operations. The case is slightly different for the thigh motor neurons. In addition to exciting 19 motor neurons in the same motor neuron pool and a V1/V2b interneuron, one thigh motor neuron also excites an IIN to inhibit the antagonistic muscle in the calf. Additionally, one thigh motor neuron in one limb sends 60 axons in total to the thigh motor neuron pools in all the other limbs. Thus, one spike from a thigh muscle motor neuron results in 81 accumulation operations. Also, when a thigh joint is approaching the limit positions, the 20 corresponding thigh muscle motor neurons receive limit position inhibitory spikes. The limit position spiking frequency here is the simulation frequency of the SNN, which is 1000Hz. Increasing the SNN simulation frequency will not influence the strength of limit position inhibition onto the motor neurons, but it will slightly raise the total power consumption. The total power consumption of the SNN-based controller is:
\begin{equation}
    \begin{split}
        P_{\mathrm{SNN-CPG}} = (&f_{\mathrm{all\_inhi}}\times 20\\
        +&f_{\mathrm{all\_calf}}\times 20\\
        +&f_{\mathrm{all\_thigh}}\times 81\\
        +&f_{\mathrm{all\_limit\_posistion}}\times 20) \times \epsilon_{\mathrm{ADD}}
    \end{split}
\end{equation}
where, $f_{\mathrm{all\_inhi}}$, $f_{\mathrm{all\_calf}}$, $f_{\mathrm{all\_thigh}}$ and $f_{\mathrm{all\_limit\_posistion}}$ are the total firing frequencies of all the inhibitory interneurons, calf muscle motor neurons, thigh muscle motor neurons and limit position inhibition. The total firing frequencies of the neurons are estimated by taking the mean of the session-wise average firing frequencies of the neurons. Specifically, the CPG runs for a session whose maximum session length is 10s to produce one session-wise average firing frequency. The estimated total firing frequency is the average of 10 session-wise average firing frequencies. The estimated average firing frequencies are $f_{\mathrm{all\_inhi}} = 6.69\times10^2\mathrm{Hz}$, $f_{\mathrm{all\_calf}} = 4.34\times10^3\mathrm{Hz}$, $f_{\mathrm{all\_thigh}} = 4.52\times10^3\mathrm{Hz}$ and $f_{\mathrm{all\_limit\_posisition}} = 2.22\times10^3\mathrm{Hz}$. Consequently, $P_{\mathrm{SNN-CPG}} = 4.60\times10^{-7}\,\,\mathrm{Watt}$, which is $23.3\times$ power efficient than the reinforcement learning based approach.

\section{Discussion}
In summary, the work proposes a bio-inspired, SNN-based control architecture for quadruped robotic systems regulated by the homeostasis property of astrocytes. The neuromorphic CPG reveals the high energy efficiency of a spike-based locomotion control system. Also, the inhibitory astrocytic regulation causing the gait emergence underscores the significant role of modeling astrocytes for neuromorphic robotic control deployed in edge robotic applications.

Future work should consider further strengthening of the neuroscience connection in CPG modeling to develop adaptive and energy-efficient robotic locomotion.
For instance, a more bio-plausible CPG modeling can consider the torque applied to a joint to have a complex relationship with the background stimulation. Multiple types of interneurons in the spinal cord, in fact, are involved in the neural circuit at different locomotion speeds \cite{kiehn2016decoding}, which could be a topic of further research. Descending command signals from the brainstem and sensory inputs are critical for animal locomotion. Also, the muscles in the torso are indispensable in multiple types of action. At first glance of animal locomotion, it seems that the legs are contributing most to the emergence of gaits. However, the limb muscles are only a fraction of the actuators of animal movement. The feed-forward plus feedback organization of all the muscles and the sensory apparatus as an entity is the true nature of animal locomotion and underlies the locomotion efficiency of biological systems.

\bibliographystyle{IEEEtran}

\section*{Acknowledgments}
We are thankful to T. Iwasaki from the Department of Mechanical and Aerospace Engineering, UCLA and R. Dzakpasu from the Department of Physics, Georgetown University for their valuable suggestions regarding the work.

Research was sponsored primarily by the Army Research Office and was accomplished under Grant Number W911NF-24-1-0127. The views and conclusions contained in this document are those of the authors and should not be interpreted as representing the official policies, either expressed or implied, of the Army Research Office or the U.S. Government. The U.S. Government is authorized to reproduce and distribute reprints for Government purposes notwithstanding any copyright notation herein. This work was also supported partially by the National Science Foundation under award No. EFRI BRAID \#2318101.

\section*{Supplementary Material}
\subsection*{Robot Simulation Model and Control Parameters}
A single limb of the A1 robot contains three joints which are called hip, thigh and calf, which can be noticed in Figure 1d. The torque outputs from the central pattern generator are directly connected to the thigh and calf joint motor actuators. The terrain is flat ground. The robot's initial heading direction is the positive $x$ direction. The thigh and calf joint limit positions are modified from the manufacturer's setting for better gait formation, and the hip joint limit position is not changed. The position range for the front and rear thigh joints are correspondingly $[\mathrm{pos^{limit,-}_{thigh,front}}, \mathrm{pos^{limit,+}_{thigh,front}}]=[0.6,1.4]$ and $[\mathrm{pos^{limit,-}_{thigh,rear}}, \mathrm{pos^{limit,+}_{thigh,rear}}]=[0.7,1.5]$. The position range for the calf joint is $[\mathrm{pos^{limit,-}_{calf}}, \mathrm{pos^{limit,+}_{calf}}]=[-1.6,-1.0]$, all position values are in unit rad. As described in the Locomotion Unit section in the main text, the inhibition region for the thigh joint is the region within 0.05 from the limit positions. The ``frictionloss'' attribute of the joints is also increased to prevent fast swing angular velocity. The ``frictionloss'' parameter for hip and calf joints is set to 10 while that for the thigh joint is set to 25. The hip joints allow the limbs to move medially or laterally. Therefore, they do not significantly contribute to the forward driving force onto the torso. Therefore, a simple PI controller is designed to stabilize the hip joints at a constant position, where $K_\mathrm{P} = 30$ and $K_\mathrm{I} = 10$. The designed stable position of the hip joints is laterally $0.1$, i.e. $0.1$ rad away from the center line of the torso. The expression of the hip joint torque as a function of the current and cumulative position error therefore is
\begin{equation}
    \begin{split}
        \mathrm{torque_{hip}}(t) = &K_{\mathrm{P}}\cdot[\mathrm{pos^*_{hip}} - \mathrm{pos_{hip}}(t)]\\
        + &K_{\mathrm{I}}\cdot\int_0^t[\mathrm{pos^*_{hip}} - \mathrm{pos_{hip}}(t)]dt
    \end{split}
\end{equation}
where, $\mathrm{pos^*_{hip}}$ is the target position of hip joints, i.e. $0.1$. $\mathrm{pos_{hip}}(t)$ is the position of hip joint at time $t$.
The resetting torso altitude is modified from the default value $0.3$ to $0.35$ to avoid conflict between the feet and the ground. 

Before the onset of each training session, $t_{\mathrm{learning\_start}}$ and 
$\mathrm{Progress}(t)$ are calculated. The hip PI controller, motor traces, MuJoCo system status and all the SNN neuron status are reset. The astrocyte status is not reset before the beginning of a new session since the $\mathrm{Ca}^{2+}$ dynamics is a slow fluctuating process that will not have an abrupt change because of the change of robot status and is a reflection of the average motor neuron activities over several seconds. 
The thigh and calf joint positions are reset to a weighted sum of the limit positions of the joints. Accurately:
\begin{equation}
    \begin{split}
        &\mathrm{pos^{reset}_{thigh,front}} = 0.7 \cdot\mathrm{pos^{limit,-}_{thigh,front}} + 0.3 \cdot\mathrm{pos^{limit,+}_{thigh,front}}\\
        &\mathrm{pos^{reset}_{rear}} = 0.7 \cdot\mathrm{pos^{limit,-}_{thigh,rear}} + 0.3 \cdot\mathrm{pos^{limit,+}_{thigh,rear}}\\
        &\mathrm{pos^{reset}_{calf}} = 0.7 \cdot\mathrm{pos^{limit,-}_{calf}} + 0.3 \cdot\mathrm{pos^{limit,+}_{calf}}
    \end{split}
\end{equation}

\subsection*{Computational Implementation of the Dynamical System}
\begin{figure}[ht]
    \centering
    \includegraphics[width=0.4\textwidth]{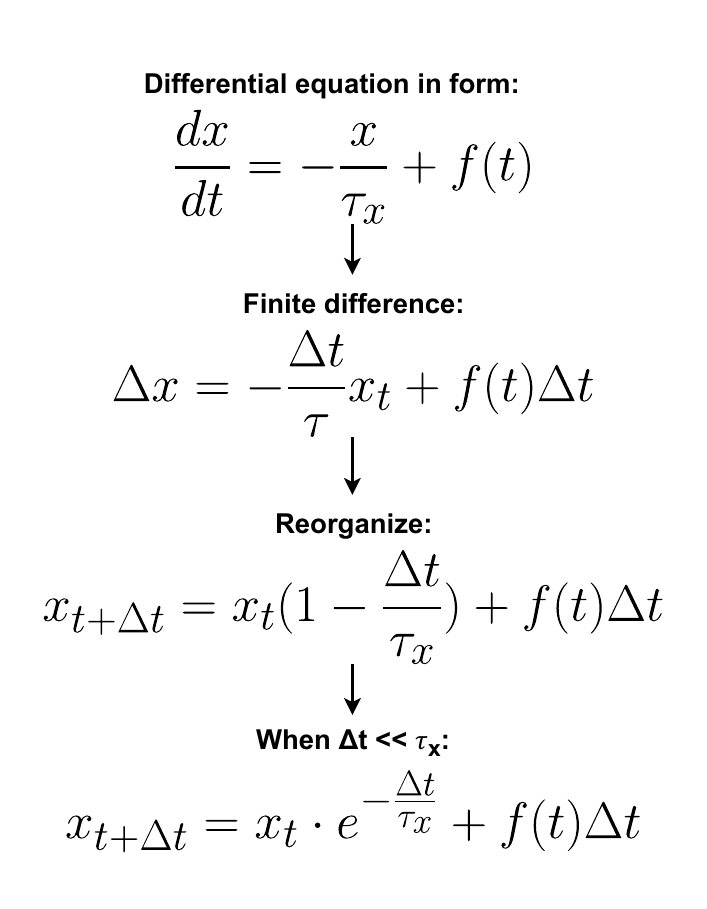}
    \caption{\textbf{Computational implementation of model dynamics.} The order of the dynamics of the models used in the paper are not higher than one. Here, $x$ represents the state variable of the system. The Euler method is adopted to simulate all the dynamics. A flowchart of derivation is shown here to illustrate the approximation used in the computation.}
    \label{computational_implementation}
\end{figure}

The STDP signal, astrocyte calcium concentration, neuron membrane potential, and the other system dynamics mentioned in this work are mostly described through a set of differential equations which have a common form. The finite difference method is adopted to implement the evolution of such dynamical systems. Figure \ref{computational_implementation} illustrates the derivation of an implementable model state updating method. For uniformity, the time constants in all the implemented dynamic systems are regarded as much greater than the simulation time step $\Delta t$.

\begin{table*}[t]

\centering
\begin{tabular}{p{4.5cm} p{9.5cm} p{2.5cm}}

Parameter & Description & Value\\
\hline

\\
\textbf{SNN-based CPG}\\
\textit{Synapse generation}\\
$w_{0\mathrm{dist}}$ & Intra-pool zero distance synapse strength & 4 \\

$c_{\mathrm{dist}}$ & Synapse strength spatial decay coefficient & 0.3 \\

$w_{\mathrm{V1/V2b}\xrightarrow{}\mathrm{motor}}$ & Uniform synapse strength from V1/V2b to motor neuron & -50 \\

$w_{\mathrm{motor}\xrightarrow{}\mathrm{V1/V2b}}$ & Uniform synapse strength from motor neuron to V1/V2b & 2 \\
\\

\textit{Motor neuron}\\
$\mathbf{Size}_\mathrm{pool}$ & Number of motor neurons in a motor neuron pool & $20$ \\

$v_{\mathrm{motor\_rest}}$ & Motor neuron resting and reseting potential & $0$ mV \\

$v_{\mathrm{motor\_th}}$ & Motor neuron threshold potential & $10$ mV \\

$\tau_{\mathrm{motor}}$ & Motor neuron membrane potential decay time constant & $9$ ms \\

$\mathrm{refractory}_{\mathrm{motor}}$ & Motor neuron refractory period & $5$ ms \\

$S_{\mathrm{motor\_spike}}$ & Motor neuron stochastic firing transition width & $0.2$ \\

$r_{\mathrm{Ca^{2+}}}$ & $[\mathrm{Ca^{2+}}]$ unit increment per action potential & $1$ \\

$\tau_{\mathrm{Ca^{2+}}}$ & $[\mathrm{Ca^{2+}}]$ decay time constant & $250$ ms \\

$\mathrm{Thres}_{\mathrm{Ca}^{2+}}$ & $[\mathrm{Ca^{2+}}]$ threshold for opening the $\mathrm{K^+}$ channel & $10$ \\

$S_{\mathrm{K\_chan}}$ & $[\mathrm{Ca^{2+}}]$ sensitivity in $\mathrm{K^+}$ channel & $10$ \\

$C_{\mathrm{K\_chan}}$ & $\mathrm{K^+}$ channel induced potential drop rate & $8000$ mV/s\\

$C_{\mathrm{limit\_position}}$ & Limit position inhibitory potential drop rate & $400$ mV/s\\

$I_{\mathrm{background},0}$ & Zero-velocity background stimulation strength & $1380$ mV/s \\

$k_{\mathrm{background},v}$ & Velocity-dependent background stimulation increment coefficient & $40$ mV/(s$\cdot$m/s) \\

$\mathrm{A}_{\mathrm{random}}$ & Randomness amplitude of background stimulation & 0.5 \\
\\
\textit{V1/V2b}\\
$v_{\mathrm{V1/V2b\_rest}}$ & V1/V2b resting and reseting potential & $0$ mV \\

$v_{\mathrm{V1/V2b\_th}}$ & V1/V2b threshold potential & $10$ mV \\

$\tau_{\mathrm{V1/V2b}}$ & V1/V2b membrane potential decay time constant & $9$ ms \\

$\mathrm{refractory}_{\mathrm{V1/V2b}}$ & V1/V2b refractory period & $3$ ms \\

$S_{\mathrm{V1/V2b\_spike}}$ & V1/V2b stochastic firing transition width & $0.2$ \\

\\
\textit{Torque output}\\
$\tau_{h}$ & Motor torque trace decay time constant & $0.1$ s \\

$I_{\mathrm{thigh}}$ & Torque increment for a single spike at thigh joint & $0.7$ $\mathrm{N}\cdot\mathrm{m}$ \\

$I_{\mathrm{calf}}$ & Torque increment for a single spike at calf joint & $1.1$ $\mathrm{N}\cdot\mathrm{m}$ \\

$\mathrm{pos^*_{hip}}$ & Hip joints' target position & lateral $0.1$ rad \\

$K_\mathrm{P}$ & Hip controller P coefficient & $30$ \\

$K_\mathrm{I}$ & Hip controller I coefficient & $10$ \\

\hline
\\

\end{tabular}
\caption{Parameters}
\end{table*}

\begin{table*}[t]
\centering
\begin{tabular}{p{4.5cm} p{9.5cm} p{2.5cm}}
Parameter & Description & Value\\
\hline
\\
\textbf{Reward-modulated STDP}\\
$[w_{\mathrm{min}}, w_{\mathrm{max}}]$ & Weight constraint factor & $[-0.05,0.05]$ \\

$\tau_{\mathrm{trace}}$ & Motor neuron spike trace decay time constant & $0.01$ s \\

$\tau_{\mathrm{STDP}}$ & STDP signal decay time constant & $2$ s \\

$\eta_{\mathrm{negative\_relative}}$ & Relative negative learning rate in STDP signal & $0.3$ \\

$\eta$ & Learning rate & $5\times 10^{-10}$ \\

$c_{\mathrm{average}}$ & Averaged reward coefficient & $0.5$ \\

$\alpha_{\mathrm{vel}_x}$ & $x$-directional velocity reward coefficient & $1$ \\

$\alpha_{\omega_{\mathrm{roll}}}$ & Roll angular velocity reward coefficient & $-0.1$ \\

$\alpha_{\omega_{\mathrm{pitch}}}$ & Pitch angular velocity reward coefficient & $-0.1$ \\

$\alpha_{\omega_{\mathrm{yaw}}}$ & Yaw angular velocity reward coefficient & $-0.1$ \\

\\
\textbf{Astrocyte dynamics}\\

$r_{\mathrm{AG}}$ & 2-AG production rate & $1\times 10^{-3}$\\

$\tau_{\mathrm{AG}}$ & [2-AG] decay time constant & $1$ s\\

$\mathrm{Thres\_Ca}_{\mathrm{ADO}}$ & $[\mathrm{Ca}^{2+}]$ threshold for adenosine release & $0.3$ \\

$r_{\mathrm{ADO}}$ & Adenosine production rate & $0.01$ \\

$\tau_{\mathrm{AG}}$ & Adenosine decay time constant & $1$ s\\

$\mathrm{refractory}_{\mathrm{ADO}}$ & Adenosine release refractory period & $0.3$ s \\

$\eta_{\mathrm{ADO}}$ & Adenosine inhibition efficacy & $1.8\times 10^{-5}$ \\
\\
\textbf{MuJoCo model}\\
$f_{\mathrm{simulation}}$ & MuJoCo simulation frequency & $1000$ Hz \\

$L_{\mathrm{max}}$ & Maximum session length & $10$ s \\

$L_\mathrm{non\_alive\_threshold}$ & Non-alive counter threshold & $0.5$ s \\

$\mathrm{Thres}_{z\_\mathrm{ALIVE}}$ & Alive torso vertical-ness threshold & $0.5$ \\

$[\mathrm{pos^{limit,-}_{thigh,front}}, \mathrm{pos^{limit,+}_{thigh,front}}]$ & Front thigh joint limit position & $[0.6,1.4]$ rad \\

$[\mathrm{pos^{limit,-}_{thigh,rear}}, \mathrm{pos^{limit,+}_{thigh,rear}}]$ & Rear thigh joint limit position & $[0.7,1.5]$ rad \\

$[\mathrm{pos^{limit,-}_{calf}}, \mathrm{pos^{limit,+}_{calf}}]$ & Calf joint limit position & $[-1.6,-1.0]$ rad \\

$\mathrm{Z_{torso,initial}}$ & Initial torso altitude & $0.35$ m \\

$\mathrm{frictionloss_{thigh}}$ & Thigh joint frictionloss & $25$ \\

$\mathrm{frictionloss_{calf}}$ & Calf joint frictionloss & $10$ \\

$\mathrm{frictionloss_{hip}}$ & Hip joint frictionloss & $10$ \\

\hline
\\

\end{tabular}
\caption{Parameters cont.}
\end{table*}

\vfill

\end{document}